\documentclass[11pt]{article}
\usepackage{graphicx}
\usepackage{amssymb}
\usepackage{setspace}
\usepackage[T1]{fontenc}
\usepackage[latin1]{inputenc}
\newtheorem{theorem}{Theorem}[section]
\newtheorem{lemma}[theorem]{Lemma}
\newtheorem{e-proposition}[theorem]{Proposition}

\newtheorem{e-definition}[theorem]{Definition\rm}

\date{}

\begin{document}

\doublespacing
\pagestyle{myheadings}
\markboth{SOM and distortion measure}{SOM and distortion measure}
\title{Self Organizing Map algorithm and distortion measure}
\author{Joseph Rynkiewicz\\SAMOS/MATISSE, Universit\'e de Paris-I,
\\ 90, rue de Tolbiac 75013 Paris, France,\\
 Tel. and Fax : (+33) 144078705\\
joseph.rynkiewicz@univ-paris1.fr}
\maketitle

\newpage
\thispagestyle{empty}
\begin{center}
{\huge Self Organizing Map algorithm and distortion measure}
\end{center}
\hrule
\begin{abstract}
% Texte de l'abstract en anglais
We study the statistical meaning of the minimization of distortion measure and the relation between the equilibrium points of the SOM algorithm and the minima of distortion measure. If we assume that the observations and the map lie in an compact Euclidean space, we prove the strong consistency of the map which almost minimizes the empirical distortion. Moreover, after calculating the derivatives of the theoretical distortion measure, we show that the points minimizing this measure and the equilibria of the Kohonen map do not match in general. We illustrate, with a simple example, how this occurs.
\end{abstract}
\hrule
\paragraph*{keywords}  Distortion measure, asymptotic convergence, consistency, Self Organizing Map, empirical processes, Glivenko-Cantelli class, uniform law of large numbers, general neighborhood function.
\newpage

\section{Introduction}
The distortion or distortion measure, is certainly the most popular criterion for assessing the quality of the classification of a Kohonen map (see Kohonen \cite{Kohonen2}). This measure yields an assessment of model properties with respect to the data and overcomes the absence of cost function in the SOM algorithm. Moreover, the SOM algorithm has been proven to be an approximation for the gradient of distortion measure (see Graepel et al.\cite{Obermayer}).

Although the Kohonen map is proven to converge sometimes on equilibria points, when the number of observations tends to  infinity, the learning dynamic cannot be described by a gradient descent of distortion measure for an infinite number of observations (see for example Erwin et al. \cite{Erwin}). Moreover, Kohonen \cite{Kohonen3} has shown in some examples for the one dimensional grid, that the model vector produced by the SOM algorithm does not exactly coincide with the optimum of distortion measure. 
This property seems to be paradoxical, on one hand SOM seems to minimize the distortion for a finite number of observations, but this behavior is no more true for the limit, i.e. an infinity of observations.

In this paper we will investigate the relationship between the SOM and distortion measure. Firstly we will prove the strong consistency of the estimator minimizing the empirical distortion. More precisely, we will prove that the maps almost minimizing the empirical distortion measure will converge almost surely to the set of maps minimizing the theoretical distortion measure. Secondly, we will calculate the derivatives of the theoretical distortion, and deduce from this calculation that the points minimizing the theoretical distortion differ generally from the equilibrium point of the SOM, whatever the dimension of the grid. Finally we will illustrate, with a simple example, why an apparent contradiction between the discrete and the continuous case occurs.

\section{Distortion measure}
We also assume in the sequel that the observations $\omega$ are independent and identically distributed (i.i.d.) and are of dimension $d$. We assume that the observations lie in an compact space, therefore, without loss of generality, they lie in the compact space $\left[0,1\right]^d$. We assume also that these observations follow the probability law $P$ having a density with respect to the Lebesgue measure of ${\mathbb R}^d$, this density is assumed to be bounded by a constant $B$. In the sequel we call centroid a vector of $\left[0,1\right]^d$ representing a class of observations $\omega$. We adopt in the sequel the notation of Cottrell et al. \cite{Cottrell}.
\begin{e-definition}
For $e\in {\mathbb N}^*$, $e\leq d$, we consider a set of units indexed by $I\subset \mathbb{Z}^{e}$ with the neighborhood
function $\Lambda $ from $I-I:=\left\{ i-j,\, i,j\in I\right\} $
to $\left[0,\, 1\right]$ satisfying $\Lambda \left(k\right)=\Lambda \left(-k\right)$
and $\Lambda \left(0\right)=1$,  note that such neighborhood function can be discrete or continuous.
\end{e-definition}
\begin{e-definition}
Note $\Vert.\Vert$ the Euclidean norm, let
\[
D^\delta_I:=\left\{x:=(x_i)_{i\in I}\in\left([0,1]^d\right)^I,\mbox{ such that }\Vert x_i-x_j\Vert\geq\delta\mbox{ if }i\neq j\right\}
\]
be the set of centroids $x_i$ separated by, at least, $\delta$. 
\end{e-definition}
\begin{e-definition}
if $x:=(x_i)_{i\in I}$ is the set of units, the Voronoi tessellation $\left(C_{i}\left(x\right)\right)_{i\in I}$
 is defined by \[
C_{i}\left(x\right):=\left\{ \omega \in \left[0,1\right]^{d}\left|\left\Vert x_{i}-\omega \right\Vert <\left\Vert x_{k}-\omega \right\Vert \mbox {\, if\, }k\neq i\right.\right\} \]
 In case of equality we assign $\omega \in C_{i}\left(x\right)$
 thanks to the lexicographical order. Conversely, the index of the Voronoi tessellation for an observation $\omega$ will be defined by
\[
C_x^{-1}(\omega)=i\in I\mbox{, if and only if }\omega\in C_i(x)
\]

\end{e-definition}

\begin{e-definition}
distortion measures the quality of a quantification with respect to the neighborhood structure. It is defined as follows:
\begin{itemize}
\item Distortion for the discrete case (empirical distortion):
We assume that the observations are in a finite set $\Omega =\left\{ \omega _{1},\cdots ,\omega _{n}\right\}$
and are uniformly distributed on this set. Then, distortion measure  is \[
V_{n}\left(x\right)=\frac{1}{2n}\sum _{i\in I}\sum _{\omega \in C_{i}\left(x\right)}\left(\sum _{j\in I}\Lambda \left(i-j\right)\left\Vert x_{j}-\omega \right\Vert ^{2}\right)\]
\item Distortion for the continuous case (theoretical distortion):
Let us assume that P is the distribution function of the observations. The theoretical distortion measure is 
\[
V\left(x\right)=\frac{1}{2}\sum _{i,j\in I}\Lambda \left(i-j\right)\int _{C_{i}\left(x\right)}\left\Vert x_{j}-\omega \right\Vert ^{2}dP\]
As mentioned before the distribution P has a density with
respect to the Lebesgue measure bounded by a constant $B>0$. 
\end{itemize}
\end{e-definition}

The distortion measure is well known to be not continuous with respect to the centroids $(x_i)_{i\in I}$ for the discrete case. Indeed, if an observation is exactly on an hyperplan separating two centroids, shifting one of the centroids will imply a jump for the distortion. So, the distortion is not continuous and, in general,  a map which realizes the minimum of the empirical distortion, does not exist. However, if we consider the sequences of maps $x^n$ such that the distortion $V_n(x^n)$ will be sufficiently close to its minimum, then we will show that such sequences of maps $x^n$ will converge almost surely to the set of maps which reaches the minimum of the theoretical distortion measure $V(x)$.
\section{Consistency of the almost minimum of distortion}
This demonstration is an extended version of Rynkiewicz \cite{Rynkiewicz}. It follows the same line as Pollard \cite{Pollard}, so  we will first show a uniform law of large numbers and then deduce the strong consistency property.
\subsection{Uniform law of large number}
Let the family of functions be
\begin{equation}
{\mathcal G}:=\left\{ g_x(\omega):=\sum_{j\in I} \Lambda\left(C_x^{-1}(\omega)-j\right)\Vert x_j-\omega\Vert^2\mbox{ for } x\in D_I^\delta\right\}
\end{equation}

In order to show the uniform law of large numbers, we have to prove that: 
\begin{equation}
\sup_{x\in D^\delta_I}\left|\int g_x(\omega)dP_n(\omega)-\int g_x(\omega)dP(\omega)\right|\stackrel{a.s.}{\stackrel{n\rightarrow\infty}{\longrightarrow}}0
\label{lgnu}\end{equation}
since, for all probability measure $Q$ on $[0,1]^d$:
\begin{equation}
\int g_x(\omega)dQ(\omega)=\int \sum_{j\in I}\Lambda\left(C_x^{-1}(\omega)-j\right)\Vert x_j-\omega \Vert^2dQ(\omega)=
\frac{1}{2}\sum_{i,j\in I}\Lambda(i-j)\int_{C_i(x)}\Vert x_j-\omega \Vert^2 dQ(\omega)
\end{equation}
Now, a sufficient condition to verify the equation (\ref{lgnu}) is the following (see Gaenssler and Stute \cite{Gaenssler}): $\forall \varepsilon > 0, \forall x_0 \in D_I^\delta$ a neighborhood $S(x_0)$ of $x_0$ exists  such that 
\begin{equation}
\int g_{x_0}(\omega)dP(\omega)-\varepsilon<\int\left(\inf_{x\in S\left(x_0\right)}g_x(\omega)\right)dP(\omega)
\leq \int \left( \sup_{x\in S(x_0)} g_x(\omega)\right)dP(\omega)<\int g_{x_0}(\omega)dP(\omega)+\varepsilon
\end{equation}
First we prove the following result, using a similar technique as the proof of lemma 11 of Fort and Pag\`es \cite{Fort}
\begin{lemma}\label{lemme1}
Let $x\in D^\delta_I$ and $\lambda$ be the Lebesgue measure on  $[0,1]^d$. Note $E^c$ the complementary set of set $E$ in $[0,1]^d$ and $|I|$ the cardinal of set I. For $0<\alpha<\frac{\delta}{2}$, let\\
\(
U^\alpha_i(x)=\left\{\omega \in [0,1]^d/\exists y\in D^\delta_I, x_j=y_j\mbox{ if } j\neq i\mbox{ and }\Vert x_i-y_i\Vert<\alpha \mbox{ and } \omega \in C_i^c(y)\cap C_i(x)\right\}
\)\\
be the set of $\omega$ changing of Voronoi cells when the  centroid $x_i$ are moving a distance of at most $\alpha$. Then
\begin{equation}
sup_{x\in D_I^\delta}\lambda\left(U^\alpha_i(x)\right)< \left(|I|-1\right)\left(\frac{2\alpha}{\delta}+\alpha\right)\left(\sqrt{2}\right)^{d-1}
\end{equation}
\end{lemma}
\paragraph{proof}

Let $x\mbox{ and } y \in D^\delta_I$ checking the assumption of  lemma \ref{lemme1} and $j\neq i \in I$. In order to prove the inequality, we have to bound the measure of $\omega$ belonging to the cells $C_i(x)$ and $C_j(y)$ simultaneously, since $\left(C_i(y)\right)^c=\bigcup_{j\in I,j\neq i}C_j(y)$.

Note $\left(z\left| t\right.\right)$, the inner product between $z$ and $t$, and
\(
\overrightarrow{n}_x^{ij}:=\frac{x_j-x_i}{\Vert x_j-x_i\Vert}
\).
The parameter vector $x+\gamma_1 \overrightarrow{n}_x^{ij}$ will be the vector with all components equal to $x$ except the component $i$ equal to $x_i+\gamma_1 \overrightarrow{n}_x^{ij}$.

Since $\Vert y_i-x_i\Vert<\alpha$, we have $\left(y_i-x_i\left| \overrightarrow{n}_x^{ij}\right.\right)=\gamma_1$ with $|\gamma_1|\leq\alpha<\frac{\delta}{2}$.
As the Lebesgue measure (of $\mathbb R^{d-1}$) of all plane sections of $[0,1]^d$ is bounded by $\left(\sqrt{2}\right)^{d-1}$, when there is a movement of the centroid $x_i$, of $\gamma_1 \overrightarrow{n}_x^{ij}$,  the Lebesgue measure of $\omega$ changing of  Voronoi cells  is then bounded by $\frac{\left|\gamma_1\right|}{2}\left(\sqrt{2}\right)^{d-1}$, so

\begin{equation}
\lambda\left(C_j\left(x+\gamma_1 \overrightarrow{n}_x^{ij}\right)\cap C_i(x)\right)<\alpha\left(\sqrt{2}\right)^{d-1}
\end{equation}
Moreover, we note that $x+\gamma_1 \overrightarrow{n}_x^{ij}$ belongs to $D_I^\frac{\delta}{2}$.

On the other hand, let $y_i-x_i-\gamma_1\overrightarrow{n}_x^{ij}:=\gamma_2 \overrightarrow{\tau}_x^{ij}$, with $\Vert \overrightarrow{\tau}_x^{ij} \Vert=1$, be the orthogonal component to $\overrightarrow{n}_x^{ij}$ of the movement of  $x_i$ to $y_i$, i.e. such that  $\left(\overrightarrow{n}_x^{ij}\left| \overrightarrow{\tau}_x^{ij}\right. \right)=0$. 

\begin{figure}[h]

\caption{\label{figureaire}hatched area $<\frac{2\gamma_2 }{\delta }<\frac{\sqrt{2}\times 2\alpha }{\delta }$}

\begin{center}\includegraphics[scale=0.5]{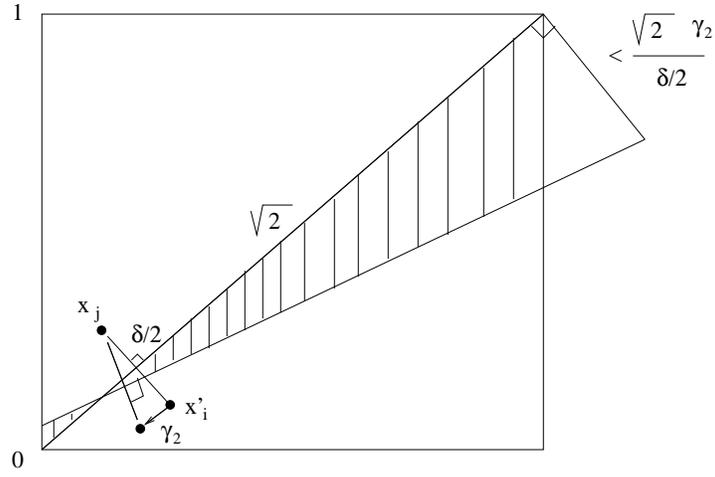}\end{center}
\end{figure}

As it is shown in figure (\ref{figureaire}), in dimension 2, for all $x'=x+\gamma_1\overrightarrow{n}_x^{ij}\in D_I^\frac{\delta}{2}$, the Lebesgue measure of $\omega$ changing of  Voronoi cells for a movement of  centroid $x'_i$, of $\gamma_2 \overrightarrow{\tau}_x^{ij}$ is bounded by $\frac{2\alpha}{\delta}\left(\sqrt{2}\right)^{d-1}$. Therefore, we have
\begin{equation}
\lambda\left(C_j\left(x+\gamma_1 \overrightarrow{n}_{x}^{ij}+\gamma_2\overrightarrow{\tau}_x^{ij}\right)\cap C_i(x)\right)< \alpha \left(\sqrt{2}\right)^{d-1}+\frac{2\alpha}{\delta}\left(\sqrt{2}\right)^{d-1}
\end{equation}

As this inequality is independent of  $x$, finally we get: 
\begin{equation}
\sup _{x\in D_I^\delta}\lambda \left( C_{j}\left( x+\gamma_1 \overrightarrow{n}_{x}^{ij}+\gamma_2 \overrightarrow{\tau }_{x}^{ij}\right)\cap C_{i}(x)\right)< \left(\alpha +\frac{2\alpha }{\delta }\right)\left(\sqrt{2}\right)^{d-1}
\end{equation}
then
\[
\sup _{x\in D_I^\delta}\lambda \left(U_{i}^{\alpha }\left(x\right)\right)< \left(\left|I\right|-1\right)\left(\alpha +\frac{2\alpha }{\delta }\right)\left(\sqrt{2}\right)^{d-1}
\]
$\blacksquare$

Now consider $x^0\in D_I^\delta$ and $S(x^0)$ a neighborhood of $x^0$ included in a sphere of radius $\alpha$. Let $W(x^0)$ be the set of $\omega$ remaining in their Voronoi cells when $x^0$ go to any $x\in S(x_0)$. For all $\omega \in W(x^0)$ we have
\begin{equation}
\begin{array}{l}
\inf_{x\in S(x^0)} g_x(\omega)\geq g_{x^0}(\omega)-\sum_{j\in I}\Lambda \left(C_{x^0}^{-1}(\omega)-j\right)\left(\Vert x^0_j-\omega \Vert^2 - \inf_{x\in S(x^0)} \Vert x^0_j-\omega \Vert^2 \right)\\
\geq g_{x_j^0}(\omega)-\sum_{j\in I}\left(\Vert x_j^0-\omega \Vert^2 - \inf_{x\in S(x^0)} \Vert x_j^0-\omega \Vert^2 \right)
\end{array}
\end{equation}
For all $\omega\in[0,1]^d$, for a small enough $\alpha$, we have 
\(
\left( \Vert x_j^0-\omega \Vert^2-\inf_{x\in S\left(x^0\right)}\Vert x_j-\omega\Vert^2\right)<\frac{\varepsilon}{2B|I|}
\) 
so
\begin{equation}
\int_{W(x^0)}\sum_{j\in I}\left(\Vert x_j^0-\omega\Vert^2-\inf_{x\in S\left(x^0\right)}\Vert x_j-\omega\Vert^2\right)dP(\omega)<\frac{\varepsilon}{2}
\mbox{ and }
\int_{W(x^0)}\left( g_{x^0}(\omega)-\inf_{x\in S(x^0)}g_x(\omega) \right)<\frac{\varepsilon}{2}
\end{equation}

Now, let $W(x^0)^c$ be the set of $\omega$ changing of Voronoi cells when the centroids go from $x^0$  to $x\in S_{x^0}$. For all $\omega\in W(x^0)^c$ there exist two different indices $i$ and $j$ such that $\omega\in C_i(x^0)$ and  $\omega\in C_j(x)$. Let us define a sequence $x^k$, $k\in\{0,\cdots,\|I|\}$,  by sequentially changing the components of $x^0$ into the components of $x$ such that $x^{|I|}=x$ ($x^k$ is the set of intermediate configurations to transform $x^0$ in $x$), then there exists a moment $l\in \{0,\cdots,|I|-1\}$, such that $\omega\in C_i(x^l)$ and $\omega\notin C_i(x^{l+1})$. Indeed, if it were not the case, you could find a sequence $x^k$, $k\in\{0,\cdots,\|I|\}$, with $x^{|I|}=x$ such that $\omega\in C_i(x^{|I|})=C_i(x)$, which would be a contradiction. So $W(x^0)^c$ is included in the set of $\omega$ which change of Voronoi set when we change sequentially the components of $x^0$ by the components of $x$.

If $\alpha<\frac{\delta}{4}$, then when the components  $x^0_i$ of $x^0$ are moving sequentially from $x^0$ to $x_i$ of $x$, each intermediate configuration stays in $D^{\frac{\delta}{2}}_I$. Since, for all $i\in I$, $\Vert x_i-\omega\Vert^2$ is bounded by $1$ on $[0,1]^d$, the lemma \ref{lemme1}, assure that
\begin{equation}
\int_{W(x^0)^c}g_x(\omega)dP(\omega)<B|I|(|I|-1))\left(\frac{4\alpha}{\delta}+\alpha\right)\left(\sqrt{2}\right)^{d-1}
\end{equation}
Finally, if we choose a small enough  $\alpha$ such that $B|I|(|I|-1))\left(\frac{4\alpha}{\delta}+\alpha\right)\left(\sqrt{2}\right)^{d-1}<\frac{\varepsilon}{2}$, we get
\begin{equation}
\int_{D_I^\delta}g_{x^0}(\omega)dP(\omega)-\varepsilon<\int_{D_I^\delta}\left(\inf_{x\in S(x^0)}g_x(\omega)\right)dP(\omega)
\end{equation}
Exactly in the same way, for a small enough $\alpha$, we get
\begin{equation}
\int_{D_I^\delta}\left(\sup_{x\in S(x^0)}g_x(\omega)\right)dP(\omega)<\int_{D_I^\delta}g_{x^0}(\omega)dP(\omega)+\varepsilon
\end{equation}
Therefore, the sufficient condition for the uniform law of large numbers is true.
\subsection{Consistency}
We want to show the consistency of the procedure involving choosing maps $(x^n)_{n\in{\mathbb N}^*}$ which almost minimizes the empirical distortions $(V_n(x))_{n\in{\mathbb N}^*}$ in $D_I^\delta$.

Let
\begin{equation}
\bar{\chi}_n^\beta:=\left\{x\in D_I^\delta\mbox{ such that } V_n(x)<\inf_{x\in D_I^\delta}V_n(x)+\frac{1}{\beta(n)}\right\}
\end{equation}
be the set of estimators that almost minimize the empirical distortion, with $\beta(n)$ being a strictly positive function, such that $\lim_{n\rightarrow+\infty}\beta(n)=\infty$.
Let
\(
\bar{\chi}=\arg\min_{x\in D_I^\delta}V(x)
\) be the set of maps minimizing the theoretical distortion, eventually reduced to one map. It is easy to verify that the function 
\(
 x\longmapsto V\left(x\right)
\)
is continuous on $D_{I}^{\delta }$, so for all neighborhood 
$\mathcal{N}$ of $\bar{\chi }$, $\eta \left(\mathcal{N}\right)>0$ exists such that
\begin{equation}
\forall x\in D_{I}^{\delta }\backslash \mathcal{N},\, V\left(x\right)>\min _{x\in D_{I}^{\delta }}V\left(x\right)+\eta \left(\mathcal{N}\right)
\end{equation}
to show the strong consistency, it is enough to prove that for all neighborhoods
$\mathcal{N}$ of $\bar{\chi }$ we have 
\begin{equation}
\lim _{n\rightarrow \infty }\bar{\chi }_{n}^{\beta }\stackrel{a.s.}{\subset }\mathcal{N}\Longleftrightarrow \lim _{n\rightarrow \infty }V\left(\bar{\chi }_{n}^{\beta }\right)-V\left(\bar{\chi }\right)\stackrel{a.s.}{\leq }\eta \left(\mathcal{N}\right)
\end{equation}
with
\(
V\left(E\right)-V\left(F\right):=\sup \left\{ V\left(x\right)-V\left(y\right)\mbox {\, for\, }x\in E\mbox {\, and\, }y\in F\right\} 
\).
By definition $V_n\left(\bar{\chi }_{n}^{\beta }\right)\stackrel{a.s.}{\leq} V_{n}\left(\bar{\chi }\right)+\frac{1}{\beta \left(n\right)}$, and the uniform law of large numbers yields 
\(
\lim _{n\rightarrow \infty }V_{n}\left(\bar{\chi }\right)-V\left(\bar{\chi }\right)\stackrel{a.s.}{=}0
\), we get then \(
\lim _{n\rightarrow \infty }V_{n}\left(\bar{\chi }_{n}^{\beta }\right)\stackrel{a.s.}{\leq }V\left(\bar{\chi }\right)+\frac{\eta \left(\mathcal{N}\right)}{2}\). 
Moreover, we have  $\lim _{n\rightarrow \infty }V\left(\bar{\chi }_{n}^{\beta }\right)-V_{n}\left(\bar{\chi }_{n}^{\beta }\right)\stackrel{a.s.}{=}0$
 and
\begin{equation}
\lim _{n\rightarrow \infty }V\left(\bar{\chi }_{n}^{\beta }\right)-\frac{\eta \left(\mathcal{N}\right)}{2}\stackrel{a.s.}{<}\lim _{n\rightarrow \infty }V_{n}\left(\bar{\chi }_{n}^{\beta }\right)\stackrel{a.s.}{\leq }V\left(\bar{\chi }\right)+\frac{\eta \left(\mathcal{N}\right)}{2}
\end{equation}
finally
\(
\lim _{n\rightarrow \infty }V\left(\bar{\chi }_{n}^{\beta }\right)-V\left(\bar{\chi }\right)\stackrel{a.s.}{\leq }\eta \left(\mathcal{N}\right)
\), this proves the strong consistency of the maps which almost minimizes the empirical distortion.

\section{Differences between the SOM algorithm and distortion measure}
Using the result of the previous section we can investigate the differences between the minima of the empirical distortion and the equilibria of the SOM algorithm. Namely, if these equilibria were maps almost minimizing the empirical distortion criterion they will converge, as the number of observations increases, to the minimum of the theoretical distortion measure but we will show that it is not generally the case. In the next section we will compute the gradient of the function $V(x)$, and show that even in multidimensional cases, the equilibria of the SOM algorithm and the minima of $V(x)$ do not match. These results generalize the results of Kohonen \cite{Kohonen3} obtained for unidimensional cases.  
\subsection{Derivability of $V\left(x\right)$}

Let us now write 
\begin{equation}
D_{I}:=\left\{ \left(x_{i}=\left(x_{i}^{1},\cdots ,x_{i}^{d}\right)\right)_{i\in I}\in \left(\left[0,1\right]^{d}\right)^{I}\left|\forall k\in \left\{ 1,\cdots ,d\right\} \left\Vert x_{i}^{k}-x_{j}^{k}\right\Vert >0\mbox {\, if\, }i\neq j\right.\right\} 
\end{equation}

For $i$ and $j\in I$, notes $\overrightarrow{n}_{x}^{ij}$ the vector
 $\frac{x_{j}-x_{i}}{\left\Vert x_{j}-x_{i}\right\Vert }$ and let
\begin{equation}
M_{x}^{ij}:=:\left\{ u\in \mathbb{R}^{d}/\left\langle u-\frac{x_{i}-x_{j}}{2},x_{i}-x_{j}\right\rangle =0\right\} 
\end{equation}
 be the mediator hyperplan. Let us note $\lambda _{x}^{ij}\left(\omega \right)$
the Lebesgue measure on $M_{x}^{ij}$. Fort and Pag\`es \cite{Fort}, have shown the following lemma: 

\begin{lemma}\label{lemme_fp}
Let $\phi $ be an $\mathbb{R}$ valued continuous function on $\left[0,1\right]^{d}$.
For $x\in D_{I}$, let be $\Phi _{i}\left(x\right):=\int _{C_{i}\left(x\right)}\phi \left(\omega \right)d\omega $.
We note also $\left(e_{1},\cdots ,e_{d}\right)$ the canonical base of
$\mathbb{R}^{d}$. The function $\Phi _{i}$ is continuously derivable
on $D_{I}$ and $\forall i\neq j,\, l\in \left\{ 1,\cdots ,d\right\} $
\begin{equation}
\frac{\partial \Phi _{i}}{\partial x_{j}^{l}}\left(x\right)=\int _{\bar{C_{i}}\left(x\right)\cap \bar{C}_{j}\left(x\right)}\phi \left(\omega \right)\left\{ \frac{1}{2}\left\langle \overrightarrow{n}_{x}^{ij},e_{l}\right\rangle +\frac{1}{\left\Vert x_{j}-x_{i}\right\Vert }\times \left\langle \left(\frac{x_{i}+x_{j}}{2}-\omega \right),e_{l}\right\rangle \right\} \lambda _{x}^{ij}\left(\omega \right)d\omega 
\end{equation}
Moreover, if we note $\frac{\partial \Phi _{i}}{\partial x_{i}}\left(x\right):=\left(\begin{array}{c}
 \frac{\partial \Phi _{i}}{\partial x_{j}^{1}}\left(x\right)\\
 \vdots \\
 \frac{\partial \Phi _{i}}{\partial x_{j}^{d}}\left(x\right)\end{array}\right)$
\begin{equation}
\frac{\partial \Phi _{i}}{\partial x_{i}}\left(x\right)=-\sum _{j\in I,j\neq i}\frac{\partial \Phi _{i}}{\partial x_{j}}\left(x\right)
\end{equation}

\end{lemma}
Then, we deduce the theorem: 

\begin{theorem}
\label{th_deriv}If $P\left(d\omega \right)=f\left(\omega \right)d\omega $,
where $f$ is continuous on $\left[0;1\right]^{d}$, then $V$ is
continuously derivable on $D_{I}$ and we have 
\begin{equation}
\begin{array}{ll}
 \frac{\partial V}{\partial x_{i}}\left(x\right) & =\sum _{k\in I}\Lambda \left(i-k\right)\int _{C_{k}\left(x\right)}\left(x_{i}-\omega \right)P\left(d\omega \right)\\
  & +\frac{1}{2}\sum _{j\in I}\sum _{k\in I,k\neq i}\left(\Lambda \left(k-j\right)-\Lambda \left(i-j\right)\right)\\
  & \times \int _{\bar{C}_{k}\left(x\right)\cap \bar{C}_{i}\left(x\right)}\left\Vert x_{j}-\omega \right\Vert ^{2}\left(\frac{1}{2}\overrightarrow{n}_{x}^{ki}+\frac{1}{\left\Vert x_{k}-x_{i}\right\Vert }\times \left(\frac{x_{i}+x_{k}}{2}-\omega \right)\right)\\
  & f\left(\omega \right)\lambda _{x}^{ki}d\omega \end{array}
\end{equation}
where $\frac{\partial V}{\partial x_{i}}\left(x\right)=\left(\begin{array}{c}
 \frac{\partial V}{\partial x_{i}^{1}}\left(x\right)\\
 \vdots \\
 \frac{\partial V}{\partial x_{i}^{d}}\left(x\right)\end{array}\right)$
\end{theorem}

\paragraph{Proof}

As the function $V\left(x\right)$ is continuous on $D_{I}$, we only have
 to show that the partial derivatives exist and are continuous.
We note $h_{i}^{l}\in \mathbb{R}^{\left|I\right|\times d}$ the vector
with all components null except the component corresponding to $x_{i}^{l}$
, which is $h>0$. Then 
\begin{equation}
\begin{array}{l}
 \frac{V\left(x+h_{i}^{l}\right)-V\left(x\right)}{h}=\\
 \frac{\frac{1}{2}\sum _{k,j\in I,\, k,j\neq i}\Lambda \left(k-j\right)\int _{C_{k}\left(x+h_{i}^{l}\right)}\left\Vert x_{j}-\omega \right\Vert ^{2}P(d\omega )-\frac{1}{2}\sum _{k,j\in I,\, k,j\neq i}\Lambda \left(k-j\right)\int _{C_{k}\left(x\right)}\left\Vert x_{j}-\omega \right\Vert ^{2}P(d\omega )}{h}\\
 +\frac{\frac{1}{2}\sum _{j\in I,\, j\neq i}\Lambda \left(i-j\right)\int _{C_{i}\left(x+h_{i}^{l}\right)}\left\Vert x_{j}-\omega \right\Vert ^{2}P(d\omega )-\frac{1}{2}\sum _{j\in I,j\neq i}\Lambda \left(i-j\right)\int _{C_{i}\left(x\right)}\left\Vert x_{j}-\omega \right\Vert ^{2}P(d\omega )}{h}\\
 +\frac{\frac{1}{2}\sum _{k\in I,k\neq i}\Lambda \left(k-i\right)\int _{C_{k}\left(x+h_{i}^{l}\right)}\left\Vert x_{i}+h_{i}^{l}-\omega \right\Vert ^{2}P(d\omega )-\int _{C_{k}\left(x\right)}\left\Vert x_{i}-\omega \right\Vert ^{2}P(d\omega )}{h}\\
 +\frac{\frac{1}{2}\left(\int _{C_{i}\left(x+h_{i}^{l}\right)}\left\Vert x_{i}+h_{i}^{l}-\omega \right\Vert ^{2}P(d\omega )-\int _{C_{i}\left(x\right)}\left\Vert x_{i}-\omega \right\Vert ^{2}P(d\omega )\right)}{h}\end{array}
\end{equation}
Where the first two lines of the sums concern centroids different from $x_i$ and the last two lines the variation involving $x_i$. Now, by applying the lemma \ref{lemme_fp}, to the first two lines of the sum we get: 

\begin{equation}
\begin{array}{l}
 \lim _{h\rightarrow 0}\frac{V\left(x+h_{i}^{l}\right)-V\left(x\right)}{h}=\frac{1}{2}\sum _{k,j\in I,\, k,j\neq i}\Lambda \left(k-j\right)\\
 \int _{\bar{C_{k}}\left(x\right)\cap \bar{C}_{i}\left(x\right)}\left\Vert x_{j}-\omega \right\Vert ^{2}\left\{ \frac{1}{2}\left\langle \overrightarrow{n}_{x}^{ki},e_{l}\right\rangle +\frac{1}{\left\Vert x_{i}-x_{k}\right\Vert }\times \left\langle \left(\frac{x_{k}+x_{i}}{2}-\omega \right),e_{l}\right\rangle \right\} \lambda _{x}^{ki}\left(\omega \right)d\omega \\
 -\frac{1}{2}\sum _{k,j\in I,\, k,j\neq i}\Lambda \left(i-j\right)\\
 \int _{\bar{C_{k}}\left(x\right)\cap \bar{C}_{i}\left(x\right)}\left\Vert x_{j}-\omega \right\Vert ^{2}\left\{ \frac{1}{2}\left\langle \overrightarrow{n}_{x}^{ki},e_{l}\right\rangle +\frac{1}{\left\Vert x_{i}-x_{k}\right\Vert }\times \left\langle \left(\frac{x_{k}+x_{i}}{2}-\omega \right),e_{l}\right\rangle \right\} \lambda _{x}^{ki}\left(\omega \right)d\omega \\
 +\lim _{h\rightarrow 0}\frac{\frac{1}{2}\sum _{k\in I,k\neq i}\Lambda \left(k-i\right)\int _{C_{k}\left(x+h_{i}^{l}\right)}\left\Vert x_{i}-\omega \right\Vert ^{2}+2h(x_{i}^{l}-w^{l})+o(h)P(d\omega )-\int _{C_{k}\left(x\right)}\left\Vert x_{i}-\omega \right\Vert ^{2}P(d\omega )}{h}\\
 +\lim _{h\rightarrow 0}\frac{\frac{1}{2}\left(\int _{C_{i}\left(x+h_{i}^{l}\right)}\left\Vert x_{i}-\omega \right\Vert ^{2}+2h(x_{i}^{l}-w^{l})+o(h)P(d\omega )-\int _{C_{i}\left(x\right)}\left\Vert x_{i}-\omega \right\Vert ^{2}P(d\omega )\right)}{h}\end{array}
\end{equation}
Then, by applying the lemma \ref{lemme_fp} to the last two lines, we get:
\begin{equation}
\begin{array}{l}
 \lim _{h\rightarrow 0}\frac{V\left(x+h_{i}^{l}\right)-V\left(x\right)}{h}=\frac{1}{2}\sum _{k,j\in I,\, k,j\neq i}\left(\Lambda \left(k-j\right)-\Lambda \left(i-j\right)\right)\\
 \int _{\bar{C_{k}}\left(x\right)\cap \bar{C}_{i}\left(x\right)}\left\Vert x_{j}-\omega \right\Vert ^{2}\left\{ \frac{1}{2}\left\langle \overrightarrow{n}_{x}^{ki},e_{l}\right\rangle +\frac{1}{\left\Vert x_{i}-x_{k}\right\Vert }\times \left\langle \left(\frac{x_{k}+x_{i}}{2}-\omega \right),e_{l}\right\rangle \right\} \lambda _{x}^{ki}\left(\omega \right)d\omega \\
 +\frac{1}{2}\sum _{k\in I,k\neq i}\Lambda \left(k-i\right)\\
 \int _{\bar{C_{k}}\left(x\right)\cap \bar{C}_{i}\left(x\right)}\left\Vert x_{i}-\omega \right\Vert ^{2}\left\{ \frac{1}{2}\left\langle \overrightarrow{n}_{x}^{ki},e_{l}\right\rangle +\frac{1}{\left\Vert x_{i}-x_{k}\right\Vert }\times \left\langle \left(\frac{x_{k}+x_{i}}{2}-\omega \right),e_{l}\right\rangle \right\} \lambda _{x}^{ki}\left(\omega \right)d\omega \\
 -\frac{1}{2}\sum _{k\in I,k\neq i}\int _{\bar{C_{k}}\left(x\right)\cap \bar{C}_{i}\left(x\right)}\left\Vert x_{i}-\omega \right\Vert ^{2}\left\{ \frac{1}{2}\left\langle \overrightarrow{n}_{x}^{ki},e_{l}\right\rangle +\frac{1}{\left\Vert x_{i}-x_{k}\right\Vert }\times \left\langle \left(\frac{x_{k}+x_{i}}{2}-\omega \right),e_{l}\right\rangle \right\} \lambda _{x}^{ki}\left(\omega \right)d\omega \\
 +\sum _{k\in I}\Lambda \left(k-i\right)\int _{C_{k}\left(x\right)}(x_{i}^{l}-w^{l})P(d\omega )\end{array}
\end{equation}
finally 
\begin{equation}
\begin{array}{l}
 \lim _{h\rightarrow 0}\frac{V\left(x+h_{i}^{l}\right)-V\left(x\right)}{h}=\frac{\partial V}{\partial x_{i}^{l}}\left(x\right)=\frac{1}{2}\sum _{k,j\in I,\, k\neq i}\left(\Lambda \left(k-j\right)-\Lambda \left(i-j\right)\right)\\
 \int _{\bar{C_{k}}\left(x\right)\cap \bar{C}_{i}\left(x\right)}\left\Vert x_{j}-\omega \right\Vert ^{2}\left\{ \frac{1}{2}\left\langle \overrightarrow{n}_{x}^{ki},e_{l}\right\rangle +\frac{1}{\left\Vert x_{i}-x_{k}\right\Vert }\times \left\langle \left(\frac{x_{k}+x_{i}}{2}-\omega \right),e_{l}\right\rangle \right\} \lambda _{x}^{ki}\left(\omega \right)d\omega \\
 +\sum _{k\in I}\Lambda \left(k-i\right)\int _{C_{k}\left(x\right)}(x_{i}^{l}-w^{l})P(d\omega )\, \blacksquare \end{array}
\end{equation}
If we assume that the minimum of distortion measure is reached
in the interior of $D_{I}$ (i.e. that no centroids collapse), we
deduce from the previous results that it does not match the equilibrium
of the Kohonen algorithm. Indeed, a point $x^{*}:=\left(x_{i}^{*}\right)_{i\in I}$
asymptotically stable for the Kohonen algorithm will verify for all
$i\in I$: 
\begin{equation}
\sum _{k\in I}\Lambda \left(i-k\right)\int _{C_{k}\left(x\right)}\left(x_{i}-\omega \right)P\left(d\omega \right)=0
\end{equation}
This equation is valid even for the batch algorithm (see Fort, Cottrell and Letr\'emy \cite{Fort2}).
 It can match with a minimum of the limit distortion only if 
\begin{equation}
\begin{array}{l}
\frac{1}{2}\sum _{j\in I}\sum _{k\in I,k\neq i}\left(\Lambda \left(k-j\right)-\Lambda \left(i-j\right)\right)\\
\times \int _{\bar{C}_{k}\left(x\right)\cap \bar{C}_{i}\left(x\right)}\left\Vert x_{j}-\omega \right\Vert ^{2}\left(\frac{1}{2}\overrightarrow{n}_{x}^{ki}+\frac{1}{\left\Vert x_{k}-x_{i}\right\Vert }\times \left(\frac{x_{i}+x_{k}}{2}-\omega \right)\right)f\left(\omega \right)\lambda _{x}^{ki}d\omega=0 \end{array}
\end{equation}
but, in general, this term is not null. 
\subsection{Example of a Kohonen string with $3$ centroids}

The previous section has shown that the minimum of distortion measure does not
match the equilibrium of the Kohonen algorithm. We will illustrate
this with a simple example. The classical explanation (see Kohonen \cite{Kohonen1})
of local potential minimization by the Kohonen algorithm is far from
being satisfactory. Actually it seems that the minima of the distortion
measure  always occur on a discontinuity point, where the function
is not derivable. 

\begin{figure}[h]

\caption{Kohonen string\label{string}}

\begin{center}\includegraphics[  scale=0.5]{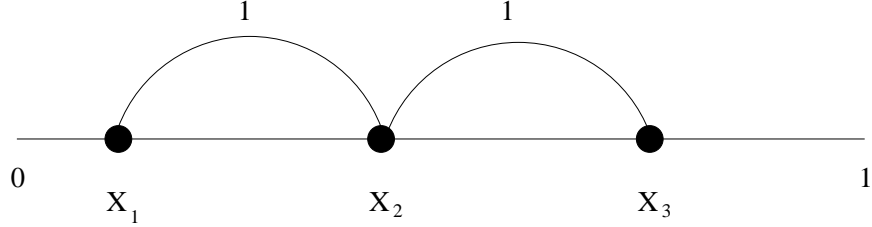}\end{center}
\end{figure}

To illustrate this, let a Kohonen string be on segment $\left[0,1\right]$
(see figure \ref{string}), with a discrete neighborhood

\subsubsection{The theoretical difference}

The equilibrium  of the SOM algorithm is reached on points $x$ verifying
\begin{equation}
\begin{array}{l}
 \frac{\partial V}{\partial x_{1}}\left(x\right)=\int _{C_{1}\left(x\right)}\left(x_{1}-\omega \right)P\left(d\omega \right)+\int _{C_{2}\left(x\right)}\left(x_{1}-\omega \right)P\left(d\omega \right)=0\\
 \frac{\partial V}{\partial x_{2}}\left(x\right)=\int _{C_{1}\left(x\right)}\left(x_{2}-\omega \right)P\left(d\omega \right)+\int _{C_{2}\left(x\right)}\left(x_{2}-\omega \right)P\left(d\omega \right)+\int _{C_{3}\left(x\right)}\left(x_{2}-\omega \right)P\left(d\omega \right)=0\\
 \frac{\partial V}{\partial x_{3}}\left(x\right)=\int _{C_{2}\left(x\right)}\left(x_{3}-\omega \right)P\left(d\omega \right)+\int _{C_{3}\left(x\right)}\left(x_{3}-\omega \right)P\left(d\omega \right)=0\end{array}
\end{equation}

but the minima of the distortion are reached on points $x$
verifying 
\begin{equation}
\begin{array}{l}
 \frac{\partial V}{\partial x_{1}}\left(x\right)=\int _{C_{1}\left(x\right)}\left(x_{1}-\omega \right)P\left(d\omega \right)+\int _{C_{2}\left(x\right)}\left(x_{1}-\omega \right)P\left(d\omega \right)-\frac{1}{4}\left\Vert x_{3}-\frac{x_{1}+x_{2}}{2}\right\Vert ^{2}f\left(\frac{x_{1}+x_{2}}{2}\right)=0\\
 \frac{\partial V}{\partial x_{2}}\left(x\right)=\int _{C_{1}\left(x\right)}\left(x_{2}-\omega \right)P\left(d\omega \right)+\int _{C_{2}\left(x\right)}\left(x_{2}-\omega \right)P\left(d\omega \right)+\int _{C_{3}\left(x\right)}\left(x_{2}-\omega \right)P\left(d\omega \right)\\
 -\frac{1}{4}\left\Vert x_{3}-\frac{x_{1}+x_{2}}{2}\right\Vert ^{2}f\left(\frac{x_{1}+x_{2}}{2}\right)+\frac{1}{4}\left\Vert x_{1}-\frac{x_{3}+x_{2}}{2}\right\Vert ^{2}f\left(\frac{x_{3}+x_{2}}{2}\right)=0\\
 \frac{\partial V}{\partial x_{3}}\left(x\right)=\int _{C_{2}\left(x\right)}\left(x_{3}-\omega \right)P\left(d\omega \right)+\int _{C_{3}\left(x\right)}\left(x_{3}-\omega \right)P\left(d\omega \right)+\frac{1}{4}\left\Vert x_{1}-\frac{x_{2}+x_{3}}{2}\right\Vert ^{2}f\left(\frac{x_{2}+x_{3}}{2}\right)=0\end{array}
\end{equation}
If we assume, for example, that the density of observations is uniform
${\mathcal{U}}_{\left[0;1\right]}$, i.e. $f(x)=1$ if $x\in \left[0;1\right]$, then these two sets of points have
no point in common. Indeed, if the two sets are equal then
\begin{equation}
\left\{
\begin{array}{l}
x_3-\frac{x_1+x_2}{2}=0\\
x_1-\frac{x_2+x_3}{2}=0
\end{array}
\right.
\end{equation}
Therefore, $x_1=x_2=x_3$, but this point is clearly not an equilibrium of the Kohonen map.

\subsubsection{Illustration of the behavior of distortion measure}

We will see that if one draws data with a uniform distribution
on the segment $\left[0,\, 1\right]$ and then one computes the minimum
of the distortion, then this minimum is always on a discontinuity point.
The more observations one has, the more  discontinuities there are,
but the global function looks more and more regular. This is not surprising,
since we know that the limit is derivable.

\paragraph{The method of simulation}

Since we have no numerical algorithm to compute the exact minimum
of variance, we proceed by exhaustive research based on a discretization
of the space of the centroids. To avoid too much computation, $0.001$ is chosen as the discretization step. 
The following figures are obtained in the following way: 

\begin{enumerate}
\item Simulate $n$ {}``data'' $\left(\omega _{1},\cdots ,\omega _{n}\right)$,
chosen with a uniform law on $\left[0,1\right]$. 
\item Search exhaustively, on the discretization of $D_{I}$, the string
which minimizes the distortion. 
\item For the best string $\left(x_{1}^{*},x_{2}^{*},x_{3}^{*}\right)$,
the graphical representations are obtained in the following way: 

\begin{itemize}
\item 3D Representation: we keep one centroid in the triplet ($x_{1}^{*}$,
$x_{2}^{*}$, $x_{3}^{*}$), then we move the other around a small
neighborhood of its optimal position. The level $z$ is the extended
variance multiplied by the number of observations $n$. 
\item 2D Representation: we keep two centroids in the triplet ($x_{1}^{*}$,
$x_{2}^{*}$, $x_{3}^{*}$), then we move the last one around a small
neighborhood of its optimal position. The level $z$ is the extended
variance multiplied by the number of observations $n$. 
\end{itemize}
\end{enumerate}
The following figures show the results obtained for a number of observations
$n$ varying from $10$, $100$ and $1000$. We notice that, even for
a small number of observations, the minima are always on discontinuity
points.

\begin{figure}[h]

\caption{Distortion measure for 10 observations}

\includegraphics[  scale=0.4,
  angle=270,
  origin=lB]{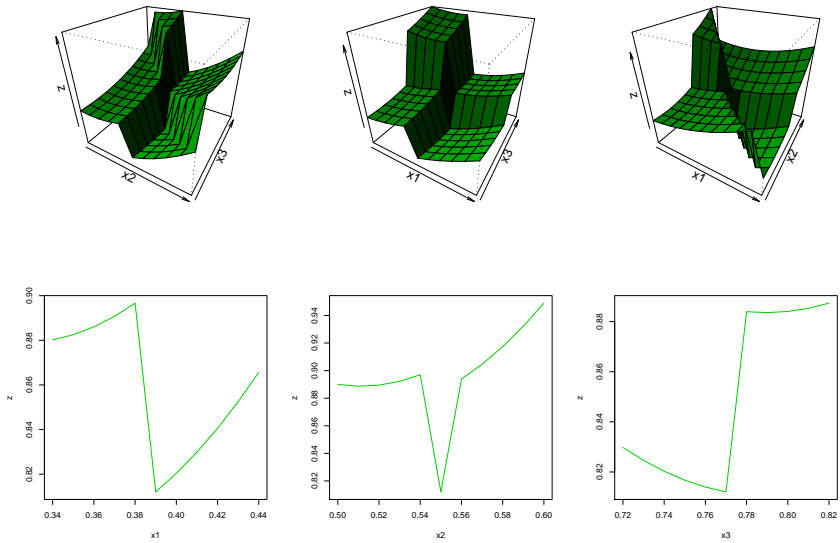}
\end{figure}

\begin{figure}[h]

\caption{Distortion measure for 100 observations}

\includegraphics[  scale=0.4,
  angle=270,
  origin=lB]{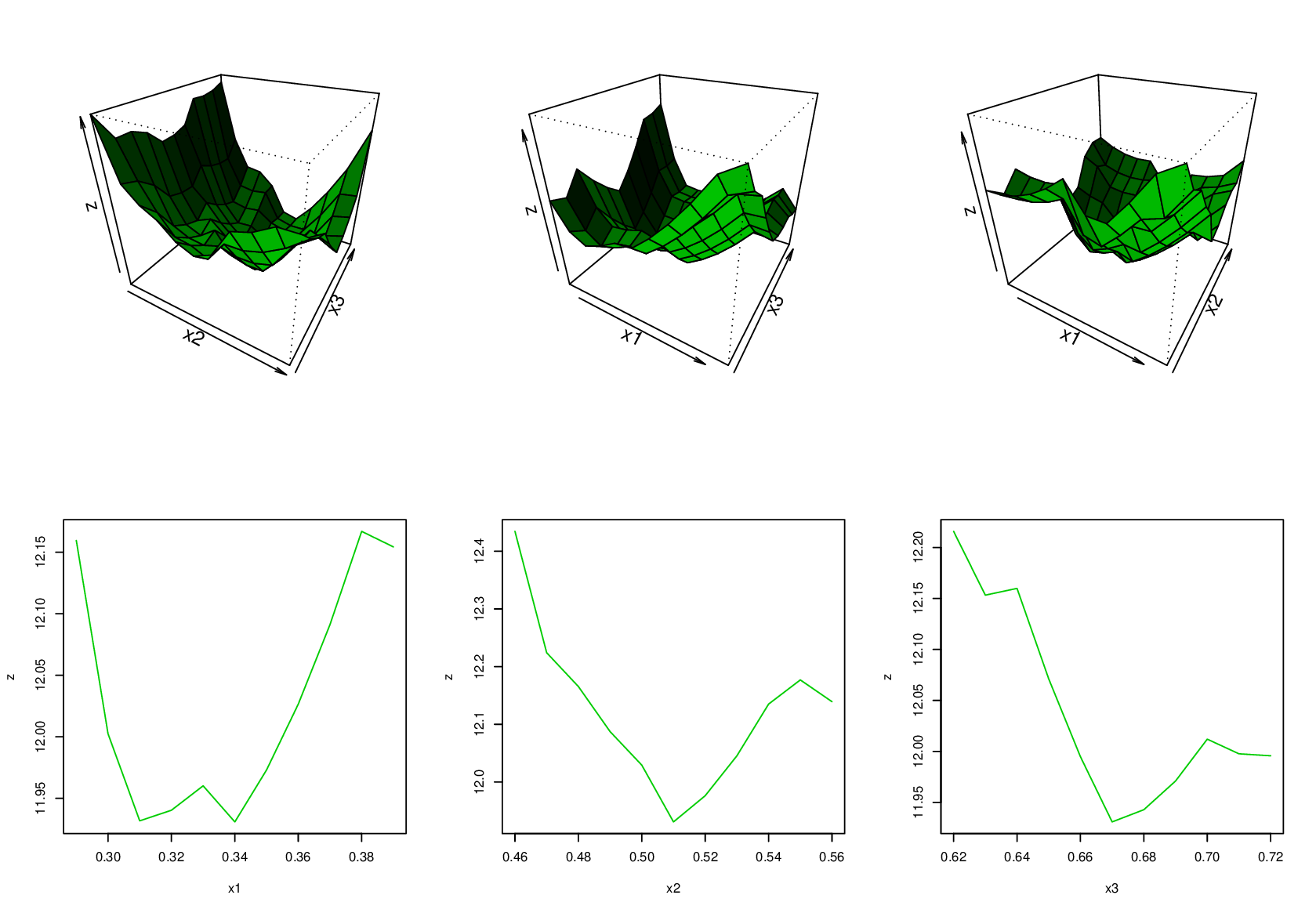}
\end{figure}
\begin{figure}[h]

\caption{Distortion measure for 1000 observations}

\includegraphics[  scale=0.4,
  angle=270,
  origin=lB]{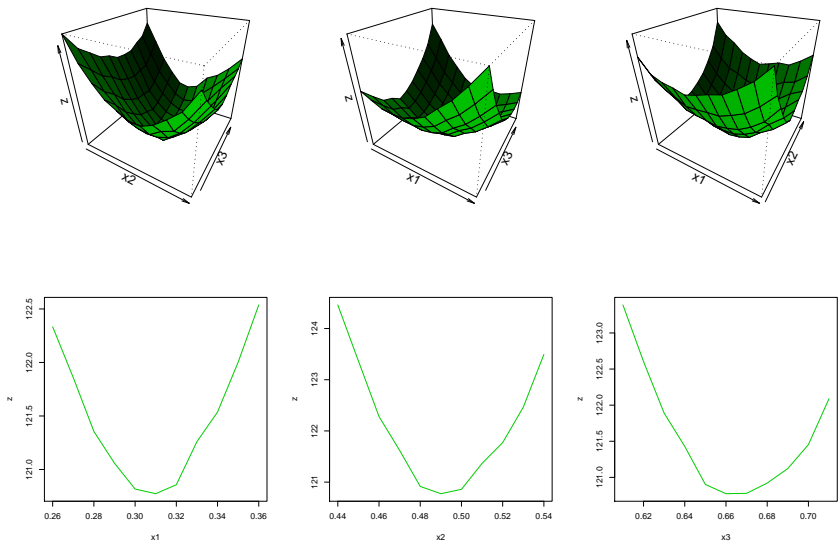}
\end{figure}

\clearpage

\section{Conclusion}

For a finite number of observations, the Kohonen algorithm was supposed to give an approximation of the
minimum of distortion measure, but if it were the case, then why can the
points of equilibrium of the algorithm be different from the
theoretical minimum of distortion? Moreover, we have shown that if we choose maps that almost minimizes the empirical distortion, then these maps have to converge to the set of maps which minimize the theoretical distortion. But, by calculating the derivative of the theoretical distortion, we have shown that the equilibria of the Kohonen map can not minimize this distortion in general. We illustrate this fact with an example where
the minimum is always reached on discontinuity points. This fact proves that the local derivability of distortion measure is not an important property and is not a satisfactory explanation for the behavior of the Kohonen algorithm when the number of observations is finite.

\end{document}